\documentclass[preprint,12pt]{elsarticle}

\usepackage{graphicx}
\usepackage{amssymb}
\usepackage{amsmath}
\usepackage{dblfloatfix}  
\journal{Journal Name}

\begin{document}

\begin{frontmatter}


\title{A Novel Indoor Positioning System for unprepared firefighting scenarios}



\author{Vamsi Karthik Vadlamani}
\author{Manish Bhattarai}
\author{Meenu Ajith}
\author{Manel Mart\'{i}nez-Ram\'on}

\address{Electrical and Computer Engineering, University of New Mexico, 1 University of New Mexico, Albuquerque, NM 87131-0001, USA.}

\begin{abstract}
 Situational awareness and Indoor location tracking for firefighters is one of the tasks with paramount importance in search and rescue operations. For Indoor Positioning systems (IPS), GPS is not the best possible solution. There are few other techniques like dead reckoning, Wifi and bluetooth based triangulation, Structure from Motion (SFM) based scene reconstruction for Indoor positioning system. However due to high temperatures, the rapidly changing environment of fires, and low parallax in the thermal images, these techniques are not suitable for relaying the necessary information in a fire fighting environment needed to increase situational awareness in real time. In fire fighting environments, thermal imaging cameras are used due to smoke and low visibility hence obtaining relative orientation from the vanishing point estimation is very difficult. The following technique that is the content of this research implements a novel optical flow  based video compass for orientation estimation and fused IMU data based activity recognition for IPS. This technique helps first responders to go into unprepared, unknown environments  and still maintain situational awareness like the orientation and, position of the victim fire fighters.
\end{abstract}


\end{frontmatter}

%\linenumbers

\section{Introduction}
\label{S:1}

In 2014, 91 fire fighter fatalities were recorded due to lack of situational aware-ness, related to position information of victims and  fire fighting officers and health monitoring. Indoor location tracking, relative orientation estimation for firefighters is a very appealing and helpful tool to reduce the first responder fatalities. The current GPS signal structure and signal power levels
are barely sufficient for indoor applications. Recent
developments in high sensitivity receiver technology are
promising for indoor positioning inside light structures
such as wooden frame houses but generally not for
concrete high rise buildings. Errors due to multipath and
noise associated with weak indoor signals limit the
accuracy and availability of global navigation satellite system (GNSS) in difficult indoor
environments \cite{5}. There are few techniques like bluetooth beacons based triangulation, wifi signal based location extraction, structure from motion based scene reconstruction and purely inertial measurement unit (IMU) based location tracking for indoor positioning system (IPS) .\newline
The WiFi based indoor localization problem (WILP)
is a difficult task in firefighting environment because the WiFi data are very noisy and highly dependent the local environmental conditions due to multi-path
and shadow fading effects in indoor environment. The 
wifi signal distribution is constantly changing depending on various
factors, such as human movement, temperature and humidity changes. In general, location-estimation systems using radio frequency (RF) signal strength function in two phases: an offline training phase and an online localization phase. In the offline phase, a radio map is built by tabulating the signal strength values received from the access points at selected locations in the area of interest. These values comprise a radio map of the physical region, which is compiled into a deterministic or statistical prediction model for the online phase. In the online localization phase, the real-time signal strength samples received from the access points are used to search the radio map to estimate the current location based
on the learned model. This simplistic approach poses a serious problem in a dynamic environment caused by the unpredictable movements \cite{21}.

Bluetooth based beacons are also proven to be very efficient in Indoor positioning systems. However, they require bluetooth beacons to be installed at specific distances for the required necessary for functionality using well-established methods techniques such as fingerprinting, trilateration, and triangulation
 with BLE beacons \cite{7,3}. In a firefighting scenario there is a high chance that the beacons might burn and hence these are not reliable methods. Similar reliability issues are associated with Wifi based IPS triangulations mentioned above.  These techniques are not suitable for a fire fighting environment due to high temperatures and constant condition fluctuations. There are also IMU based techniques which address the IPS problems. However, the use of a strapdown inertial
navigation system (INS) system and its traditional
mechanization as a personal indoor positioning system is
rather unrealistic due to the rapidly growing positioning
errors caused by IMU drifts \cite{5}. Even a high performance
INS will cause hundreds of metres of positioning error in
30 minutes without GPS updates \cite{5}. Using a standard Scene From Motion SFM technique for the scene reconstruction and camera point estimation needs to have parallax in images. Parallax is difference in position of object due to viewing object from different line of sight. It is highly impossible in thermal images because of the cameras low resolution and very less dynamic range.

The orientation information alone is a powerful tool for the firefighters as their rescue protocol and successful navigation in and out of the building  depends needs the recollection of their orientation. However, orientation in heavy smoke, heat, and other hazardous conditions associated with such environments can become extremely really difficult for them to keep track off. We  gained a deeper appreciation of the significance of maintaining a sense of  relative orientation to a first responders ability to successfully navigate when we first visited first responders training facility for data collection. There are few existing techniques to estimate relative orientation using the camera input \cite{10,12}. In this techniques they are trying to extract the relative orientation of the camera using a vanishing point estimation technique.  A vanishing point is a point on the image plane where projections of lines in image in 3D space appear to converge. If a significant number
of the imaged line segments meet very accurately in a
point, this point is very likely to be a good candidate for
a real vanishing point \cite{10}. For vanishing point estimation commonly canny edge detector \cite{11} is used followed by a RANSAC \cite{24} to remove the outliers and extract the lines that are required to estimate vanishing points. However these techniques cannot be applied to firefighting scenario because of the lack of edges that can be discerned in thermal imagery. 

For indoor positioning system, orientation estimation GPS is not a good solution because of its restricted abilities in locating indoor subjects. Beacons, WILP also does not work well due to the high temperatures and constant changes in the environment. IMU based techniques are not reliable because the errors accumulate exponentially over a period of time. It is highly difficult to obtain relative orientation using vanishing point detection due to lack of edges and quality resolution in thermal images.

To circumvent all these limitations,  we employed a novel technique to extract relative orientation estimation and devise an IPS algorithm on top of orientation estimation. 

We estimate the camera movement, obtain the relative orientation estimation using algorithms like  scale-invariant feature transform (SIFT)  \cite{25} for feature extraction and L-K optical flow \cite{20} for flow estimation. We then assign a velocity to the extracted orientations to obtain the indoor location of firefighters.

\section{Methodology}

We implement a series of individual algorithms each solving its own problem in the context of breaking down the environmental data received from the various sensors. These individual algorithms then “cascade” into one- another and their individual products are added together to recreate the entire system. Hence, we intend to introduce the flow of our system which can be observed in fig 1. \newline

 We extract SIFT features \cite{25} and employ an L-K optical flow algorithm \cite{20} on the key point positions to estimate the movement of camera. This camera movement is subject to orientation estimation routine to extract orientations. Combining the data of orientations from the video compass designed and assigning the velocity to the tensors obtained to estimate the position completes our IPS system. We primarily divide the algorithm into four important modules and will discuss about each module in following sections.

\begin{figure}
\begin{center}
    \includegraphics[width= .4\textwidth]{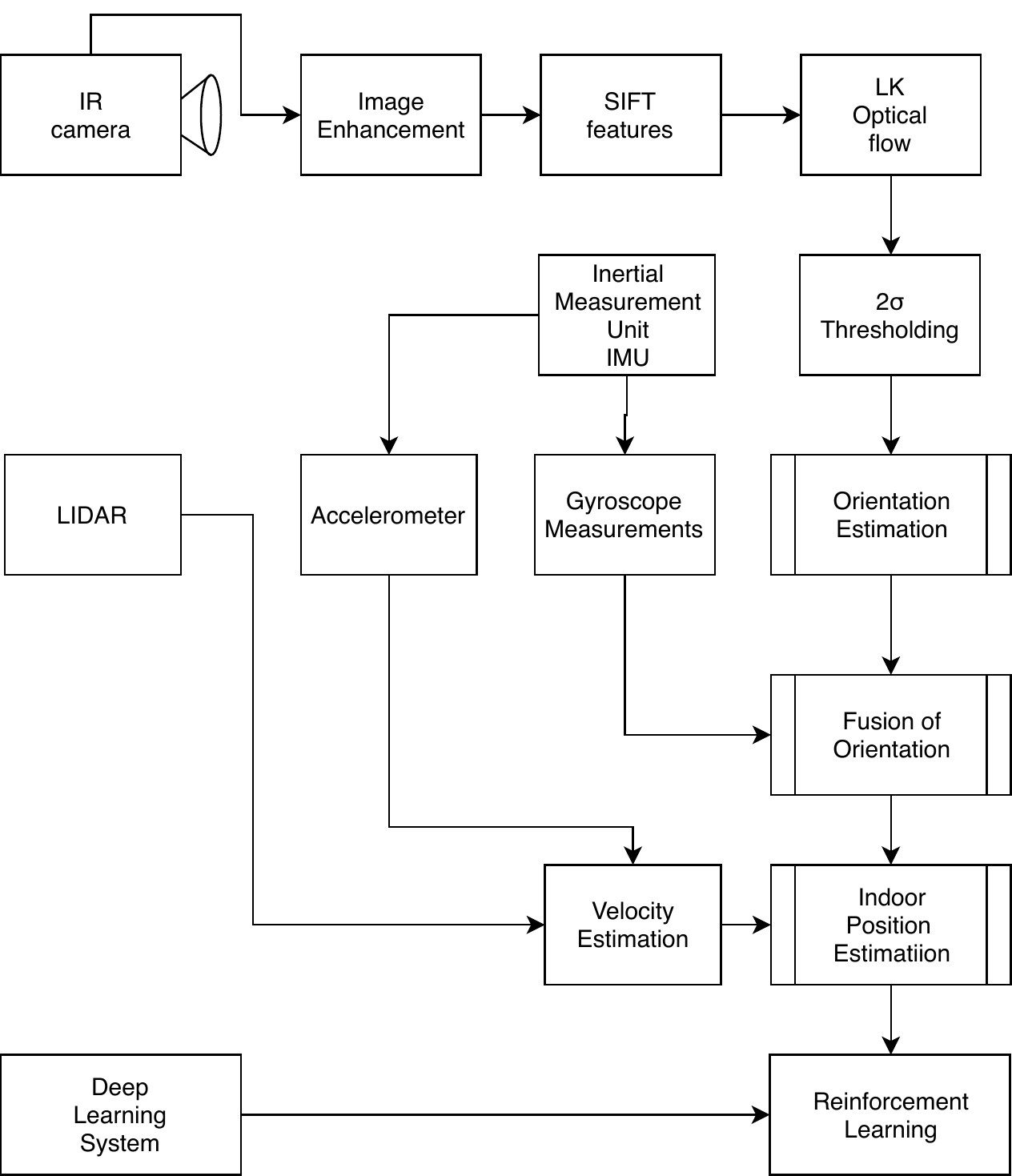}\\
    \caption{The IPS algorithm divided into sub blocks}
    \label{ActionButton}
\end{center}
\end{figure}
\subsection{Image enhancement \& Feature Extraction}

Thermal and Infrared (IR) imagery do not provide the same visual quality in terms of sharpness or level of detail needed for individual object delineation, as the RGB images that are obtained from the normal pin-hole cameras. Intense shifts in movement and orientation of the view in hand- held cameras related to the real time movements and reactions of the 
fire fighter holding the camera requires an algorithm that is robust against variability in illumination, scale and rotation. 

All the above mentioned conditions can be managed by using SIFT feature detection\cite{25} in the relative orientation estimation process.

SIFT algorithm is distillation of pixels that are passed through four stages of distillation like scale space extrema detection, keypoint localization, orientation assignment, keypoint descriptor. After each stage the pixels that are distilled will become invariant to scale, space (location), orientation. We have applied a image enhancement stage before passing it over to the strategic SIFT feature extraction stage. We have applied standard adaptive threshold processing available from an image processing library OpenCV-2.4.9.

If two subsequent images that have approximately the same content, the SIFT descriptors for these images will be similar even if one image has been changed in scale, orientation, rotation with respect to the other one. In our methodology we extract the SIFT keypoints of the frames as we assume that the keypoints of two subsequent images will be approximately the same.

The key point positions are passed over to the next stage, the Lucas-Kanade(L-K) based optical flow estimation \cite{20} together with the frames.

\subsection{Optical Flow Estimation \& Foreground activity rejection}
Optical flow in images is a measure of displacement of objects or pixels in an image considering a time sequence. Idea behind the optical flow is brightness constancy constraint. It says that a pixel value does not change in immediate frames if it can be expressed as

\begin{equation}\label{eq:optical_flow}
\begin{split}
I(x,y,t)=I(x+\Delta x,y+\Delta y,t+\Delta t)\\
U= \Delta x/ \Delta t, V= \Delta y/ \Delta t
\end{split}
\end{equation}
where x,y are pixel positions and t is time, likewise $\Delta$ represents the change in respective quantities. Also U, V are horizontal and vertical velocities respectively, they are also called flow vectors in some applications. Image $I(x,y,t)=I(x+\Delta x,y+\Delta y,t+\Delta t)$ can be approximated by a first order Taylor expansion as
\begin{equation}
\begin{split}
&I(x+\Delta x,y+\Delta y,t+\Delta t)\approx I(x,y,t)+\\&\frac{\delta I(x,y,t)}{\delta x}\Delta x
+\frac{\delta I(x,y,t)}{\delta y}\Delta y+\frac{\delta I(x,y,t)}{\delta y}\Delta t\\
&=I(x,y,t)+I_x\Delta x
+I_y\Delta y+I_t\Delta t\\
\end{split}
\end{equation}
Plugging this expression in \eqref{eq:optical_flow} and dividing by $\Delta t$ gives
\begin{equation}
I_x U+ I_yV+I_t=0\\
\end{equation}
where $I_x$, $I_y$ and $I_t$ can be approximated by finite difference computation. The L-K optical flow \cite{20} assumes that $U,V$ are constant in a small window around a pixel $I(x,y,t)$.  Then, a system of equations can be solved using Least Squares to estimate $U$ and $V$. 

Usually, this algorithm is applied to areas of the images where corners are detected. However, IR images use to lack of well defined corners. Therefore, we apply the L-K algorithm to the SIFT keypoint positions provided by the previous stage. These are the points more likely to satisfy the brightness constancy constraint in \eqref{eq:optical_flow}.

From the L-k optical flow of the matched key points, the relative movement of camera with the previous frame is estimated and we obtain velocities of the horizontal $U$ and vertical $V$ movement of camera. The firefighting scenario is highly dynamic with rapidly changing environmental conditions. The imagery that is captured in real time by first responders holds the potential for imperfect orientation estimation due to high levels of foreground movements being captured. To correct for these high levels of foreground movement, we apply a thresholding to reject higher velocities. We have observed the histogram of the data with foreground object to threshold the foreground activity.

\begin{figure}
\begin{center}
    \includegraphics[width= .4\textwidth]{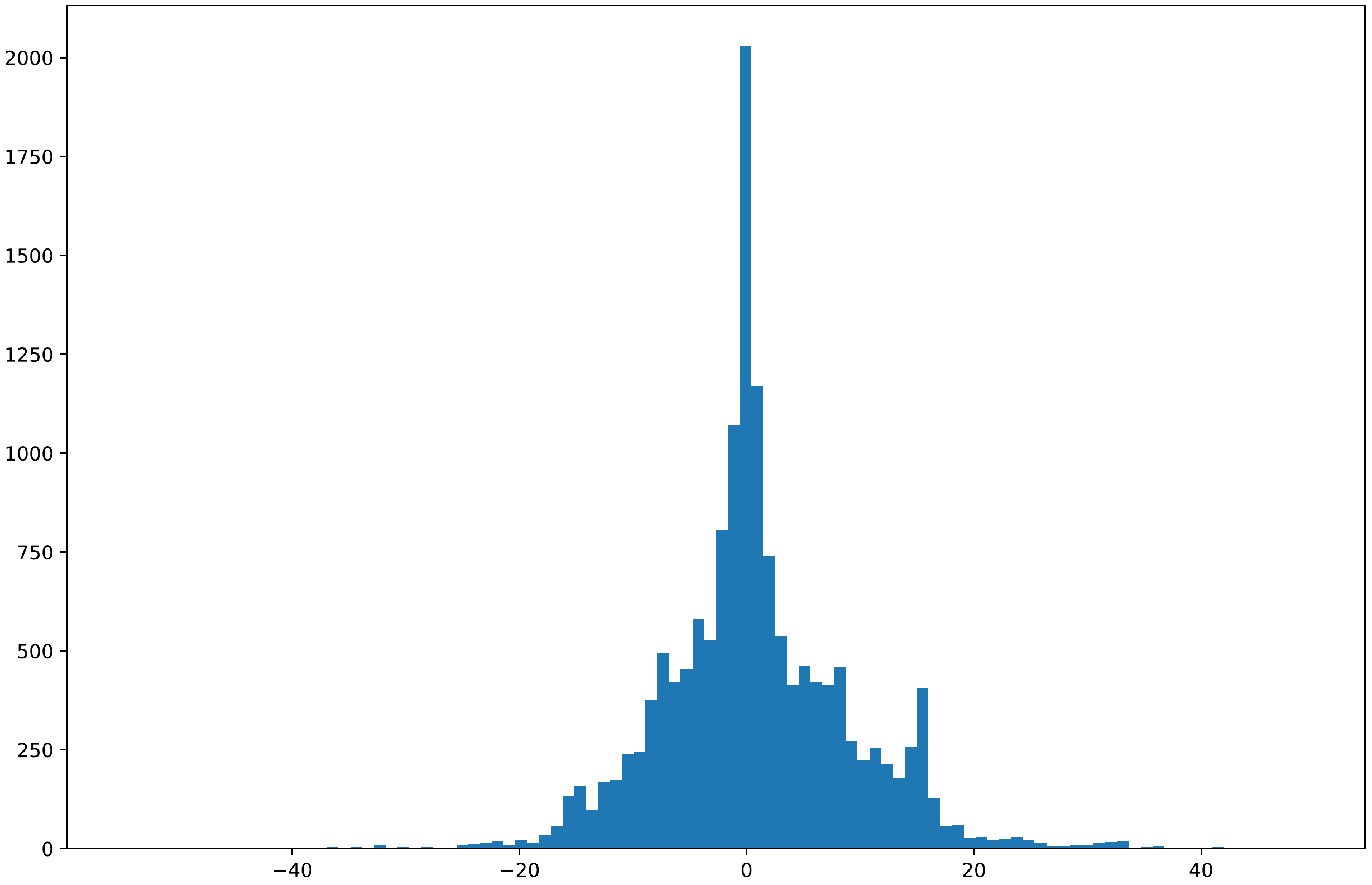}\\
    \caption{ Observed histogram for obtained velocities with foreground activity }
    \label{fig:Histogram}
\end{center}
\end{figure}
We observed that the velocities lie between 0.2 pixels/frame to 20 pixels/frame during 98\% of the time. The corresponding histogram for a set of representative frames is shown in Figure \ref{fig:Histogram}. The central peak of velocities that are less than 0.2 pixels/frame are discarded and then the standard deviation $\sigma$ is estimated for the rest of data, obtaining a  value of $\sigma = 8.39$. A threshold is applied to discard all velocities over $2\sigma \approx 17$ pixels/frame.

After thresholding, we compute the mean of the remaining flow vectors, which gives the average movement of camera. Only the horizontal velocity is of interest.

\subsection{ Relative orientation estimation}

\begin{figure}
\begin{center}
    \includegraphics[width= .5\textwidth]{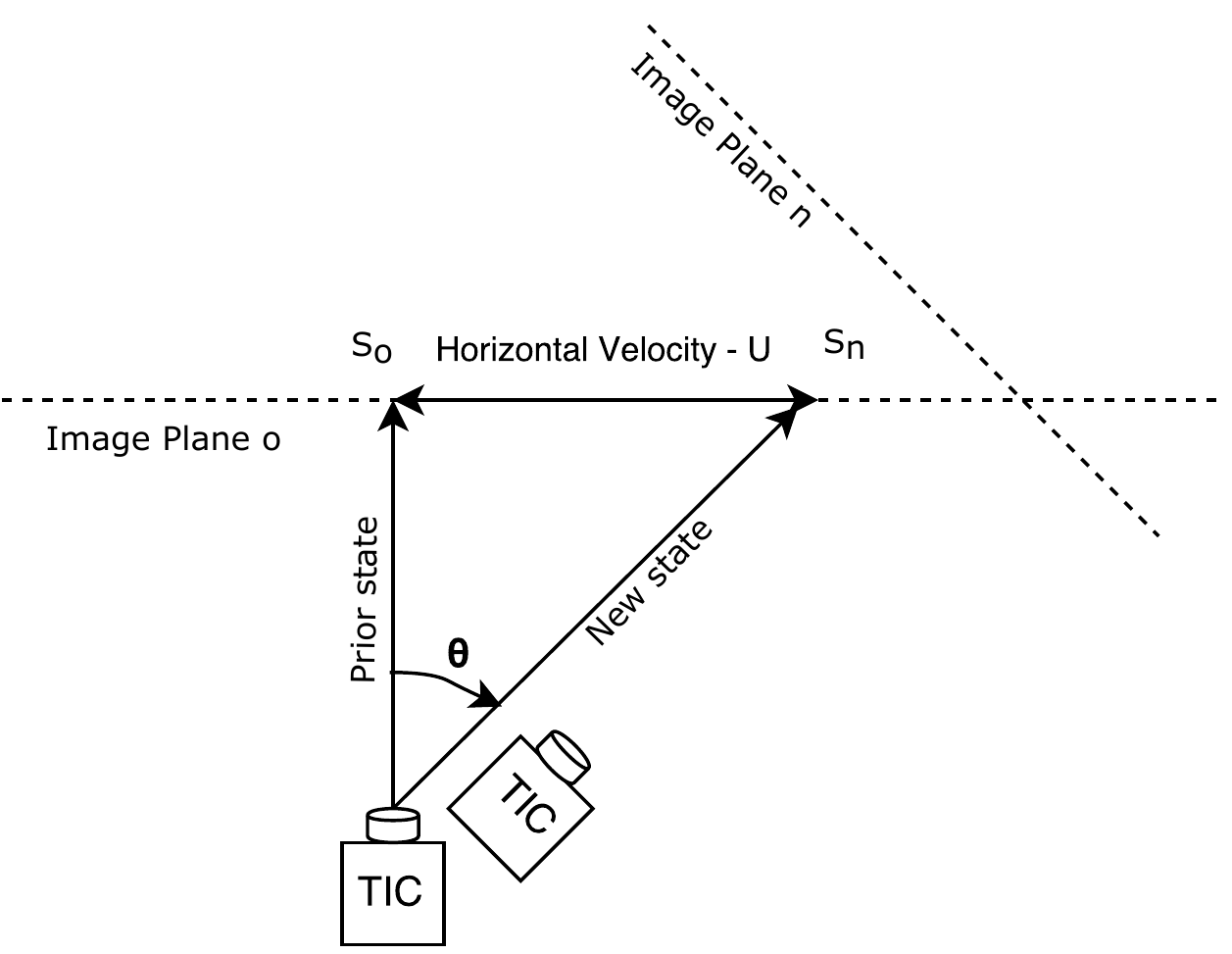}\\
    \caption{Idea of orientation estimation}
    \label{fig:Orientation_estimation}
\end{center}
\end{figure}

The estimated velocities are here used to compute the change of angle $\theta$ in radians per frame in the horizontal plane of the camera movement to estimate relative orientation of the firefighter. Figure \ref{fig:Orientation_estimation} shows the model for this estimation.

Consider the initial position of camera from which we extract SIFT keypoint positions $S_o$ of the corresponding image plane $P_o$ . When there is a change of angle $\theta$  the camera is in a  new position whose corresponding image plane $P_n$ that contains SIFT keypoint position $S_n$. To estimate $\theta$ we need to know the horizontal velocity $U$ and the distance between the camera position and image plane $P_o$. We get the horizontal velocity $U$ information from the L-K algorithm and we assign a distance $D$ from the camera to the image plane. The camera has a FOV $\beta$ and a horizontal resolution of $N_x$ pixels. If the FOV is much smaller than $180^o$, the angular resolution can be approximated by a constant $R_x=\frac{\beta}{N_x}$ degrees per pixel. This data is used to convert velocity $U$ to angular velocity. Here we consider $U$  as a tangential velocity of the camera rotation in the horizontal plane, and the corresponding angular velocity is then 
\begin{equation}
\theta=\frac{U \beta}{N_x}
\end{equation}
\begin{figure}
\begin{center}
    \includegraphics[width= 0.4 \textwidth]{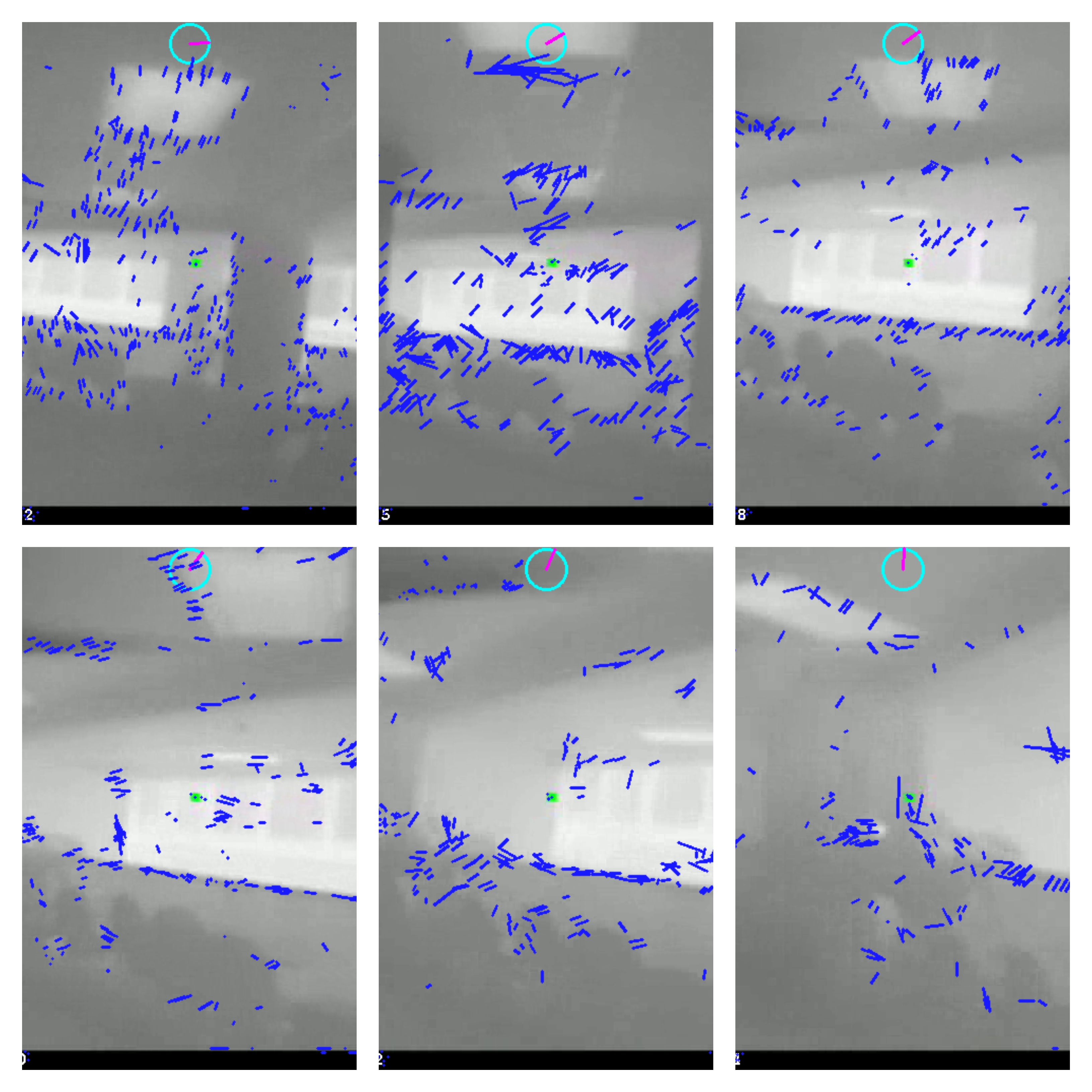}\\
    \caption{Output of orientation estimation algorithm applied to a camera sequence recorded.}
    \label{Orientation estimation}
\end{center}
\end{figure}
which will be expressed in degrees per frame. The angular velocity is then integrated across frames to compute the current orientation. This is represented in the screen as a compass as seen in figure \ref{Orientation estimation}

\begin{figure}
\begin{center}
    \includegraphics[width= 0.4 \textwidth]{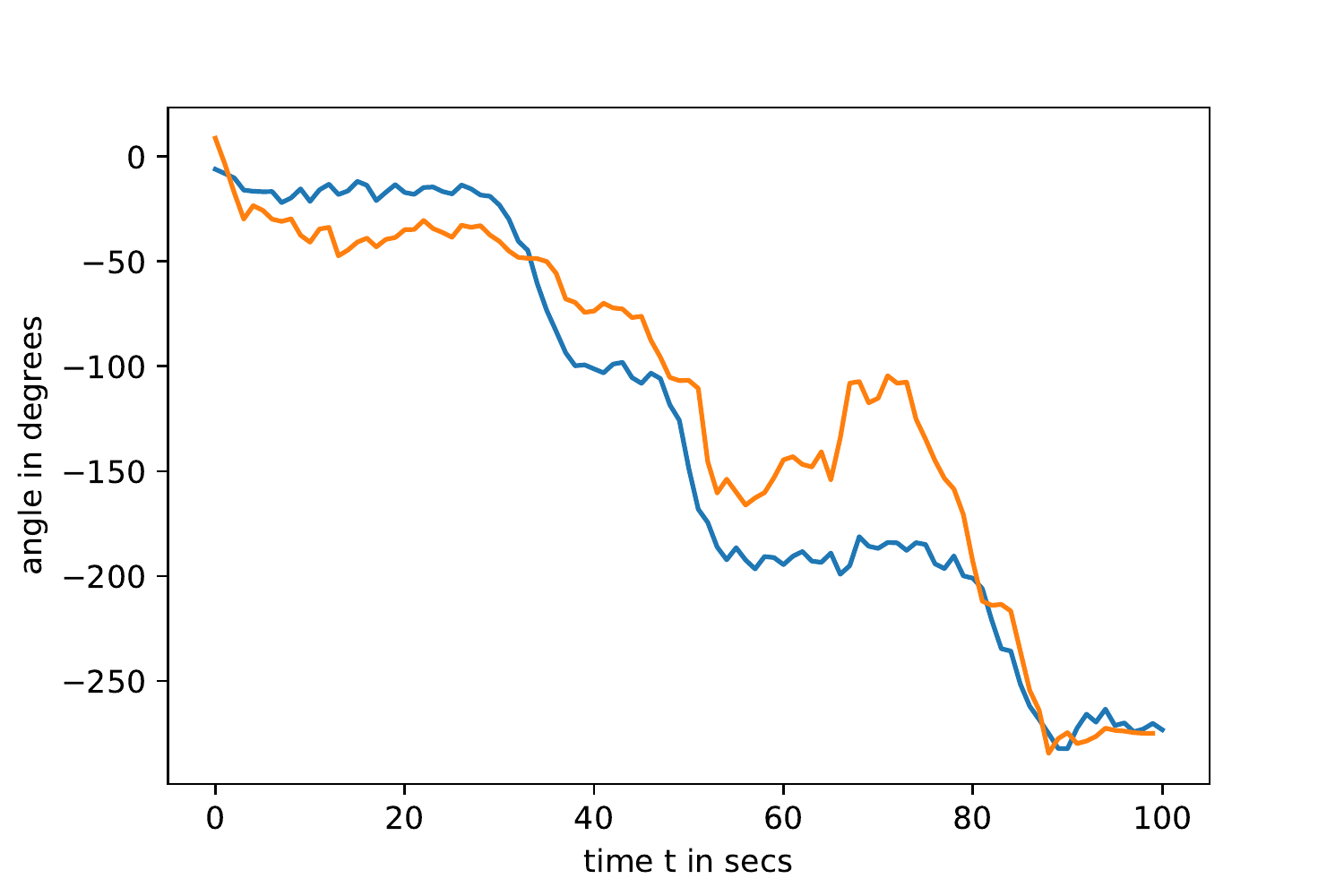}\\
    \caption{Output of orientation estimation algorithm from Camera and Gyroscope. Orange- Camera based, Blue- Gyro based}
    \label{gyro-cam}
\end{center}
\end{figure}

We have also extracted orientation information from gyroscope and have fused it with the orientation information from the relative orientation estimation based on camera feed. We can see that the orientation estimation obtained from camera and gyroscope are very similar from figure \ref{gyro-cam}. 

We fused the data from camera feed and gyroscope using a weighted summation approach. The relative orientation can be currently obtained by a simple fusion of the two information sources as follows:

\begin{equation}
\Theta= \lambda\theta_g + (1-\lambda)\theta_c
\end{equation}
Where lambda is any rational number ranging from 0 to 1. In our experiment we used $\lambda$=0.6, However we are intending to further develop this fusion process to use the information efficiently and obtain absolute robustness. In the following figure \ref{fusion_angle} we can see the fusion of the data using this two sources. 
\begin{figure}
\begin{center}
\includegraphics[width=0.4 \textwidth]{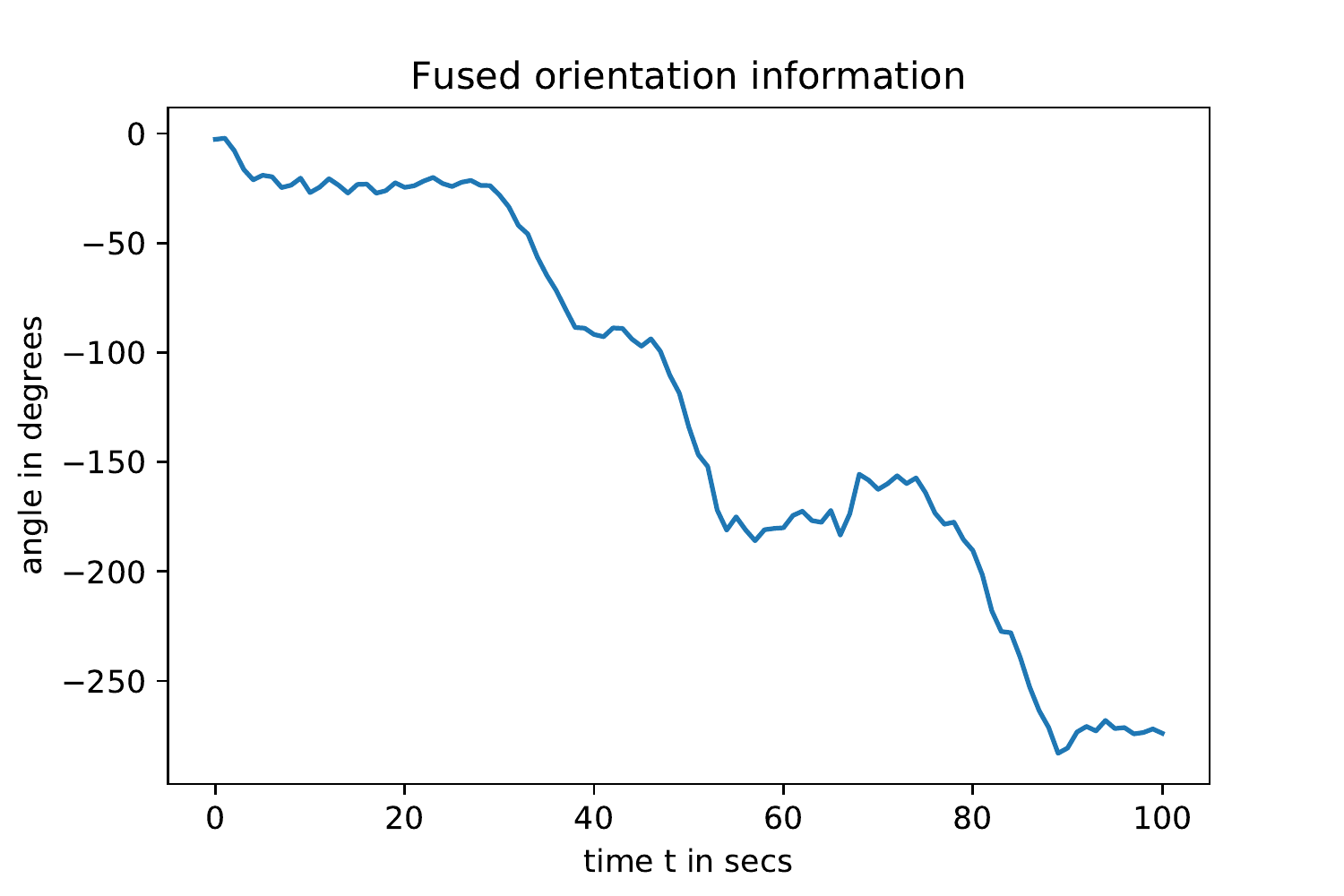}
\caption{Fusion of orientation information from Gyro and Camera based relative orientation estimation}
\label{fusion_angle}
\end{center}
\end{figure}

\subsection{LIDAR Position estimation}
Velocity estimation is a very important step for the IPS routine. To achieve this task we use the data from two sensors accelerometer and a LIDAR. We are using Garmin LIDAR v3 and interface it to an ardiuno uno board for data collection, the data is sampled at an average of 60 samples/Sec with uneven sampling which results in noise that is non zero mean, not independent and identically distributed (i.i.d). We also used accelerometer data collected at 100 hz from an IMU using Bosch XDK110 and applied a least squares estimation to fuse the accelerometer and Lidar data.

\begin{figure}[!b]
    \centering
%\begin{center}
\includegraphics[width=0.4\textwidth]{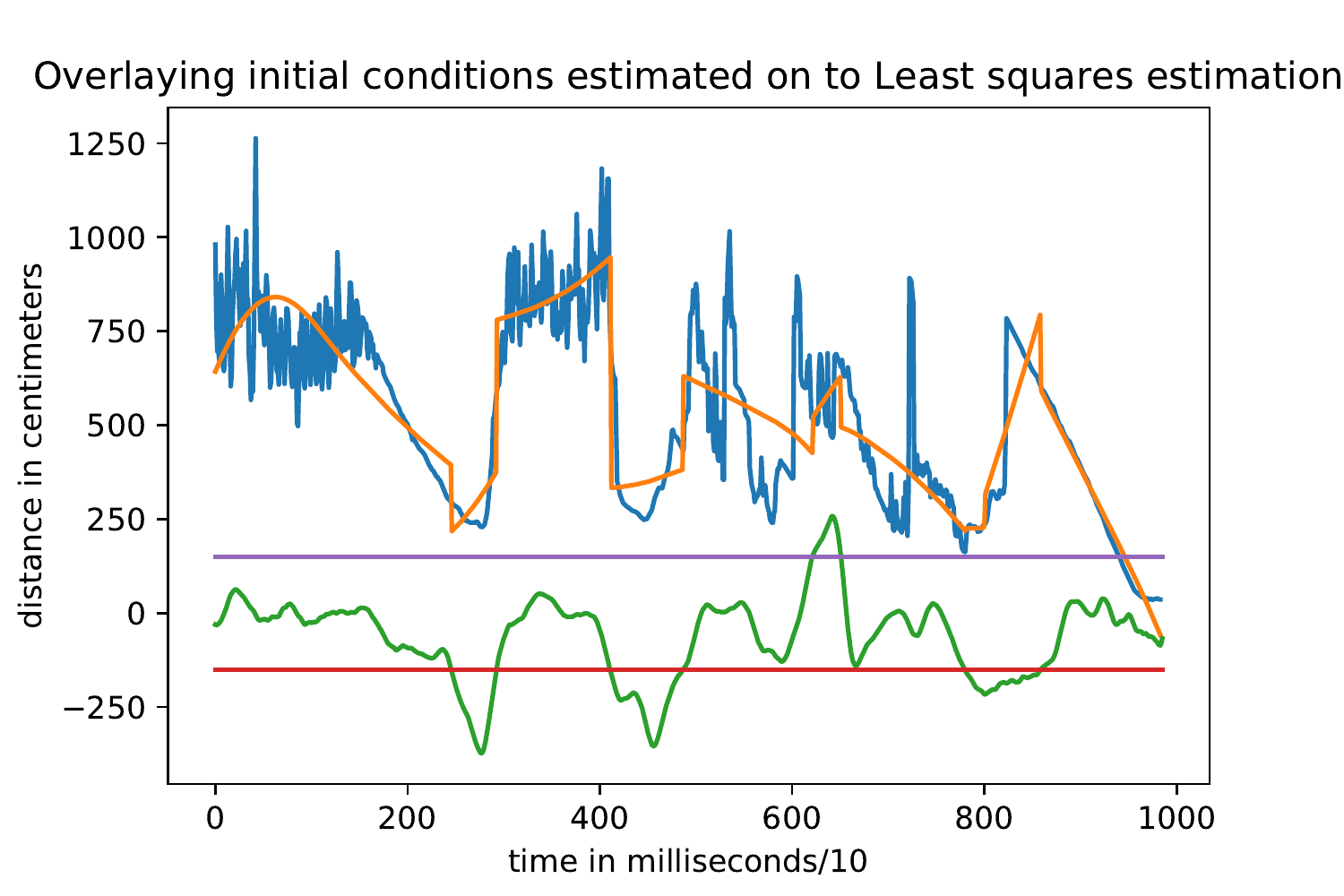}
\caption{Overlaying Least squares estimation and initial conditions to observe the raise in distance corresponding to detected turn}
\label{ini_over}
%\end{center}
\end{figure}

We employed an initial condition estimation technique to obtain initial conditions for Lidar because for every turn we make the distance estimated by lidar will increase rapidly and does not obey the laws of physics. To overcome this problem we used the orientation data and convoluted with sigmoid function. the resulted function is thresholded for peaks more than a limit and is accounted as a turn. For every turn the positions data estimation algorithms are freshly applied.  In the figure \ref{ini_over} we can see the turns detected during sudden raise of distance in the LIDAR data.

\begin{figure}
\begin{center}
\includegraphics[width=0.4 \textwidth]{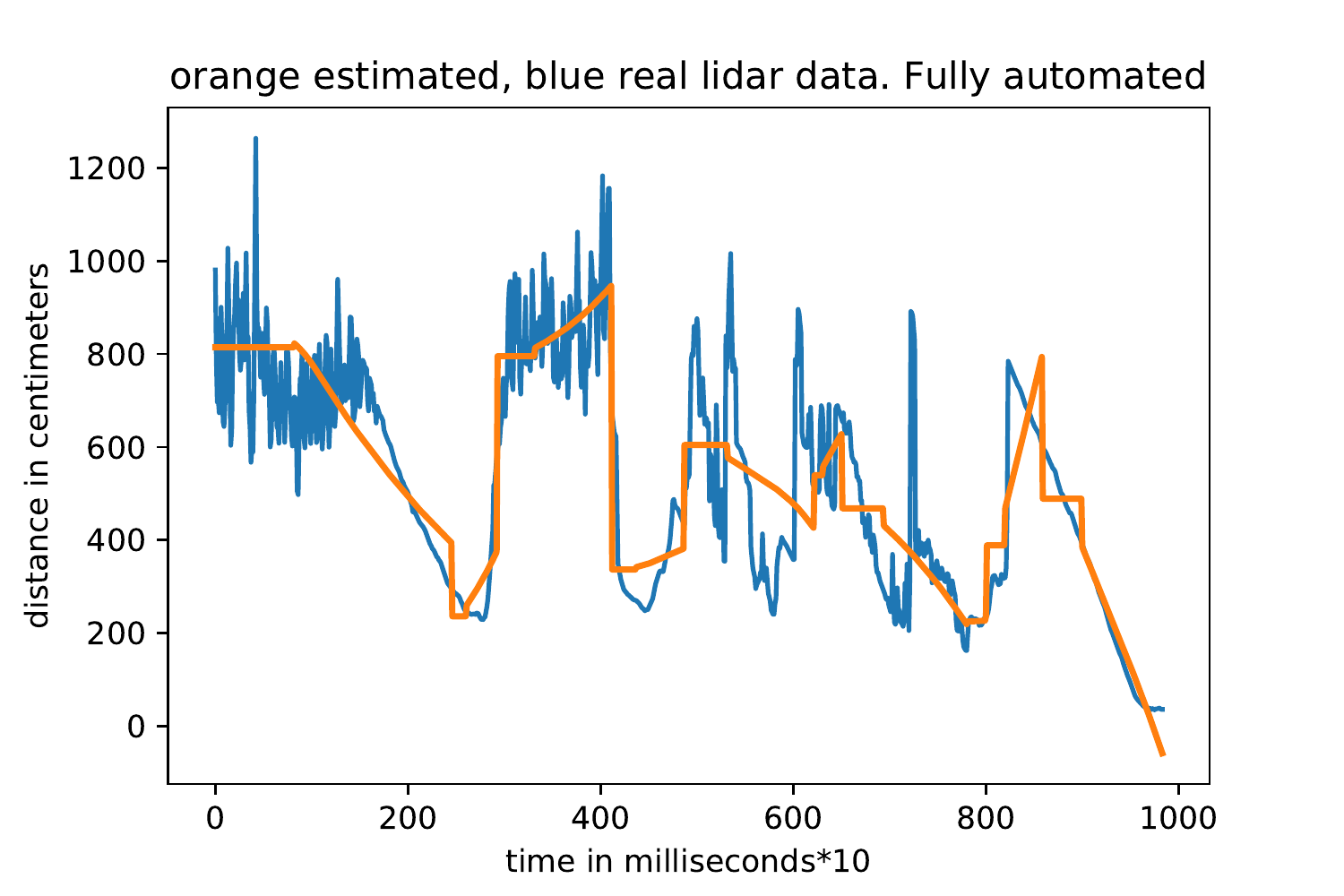}
\label{svr}
\end{center}
\caption{Position estimation using SVR}
\end{figure}

We also applied Support Vector Regression based approach to obtain the best fit of Lidar data. This approach has the advantage of having excellent generalization properties under low sample sets and of being robust to outliers thanks to its implicit cost function, which is a combination of the original Vapnik's or $\varepsilon$ cost function and the Huber robust cost function \cite{Rojo04a,martinez18}. We can see the results obtained by applying SVR in figure \ref{svr}

By applying the differentiation to the data we estimated from SVR $S_{SVR}$ we obtain the velocity information. Since there are sudden raises or falls in $S_{SVR}$, We apply a thresholding to velocities obtained $V_{SVR}$ in a way that the process obeys laws of physics. 

\begin{figure}
\begin{center}
\includegraphics[width=0.4\textwidth]{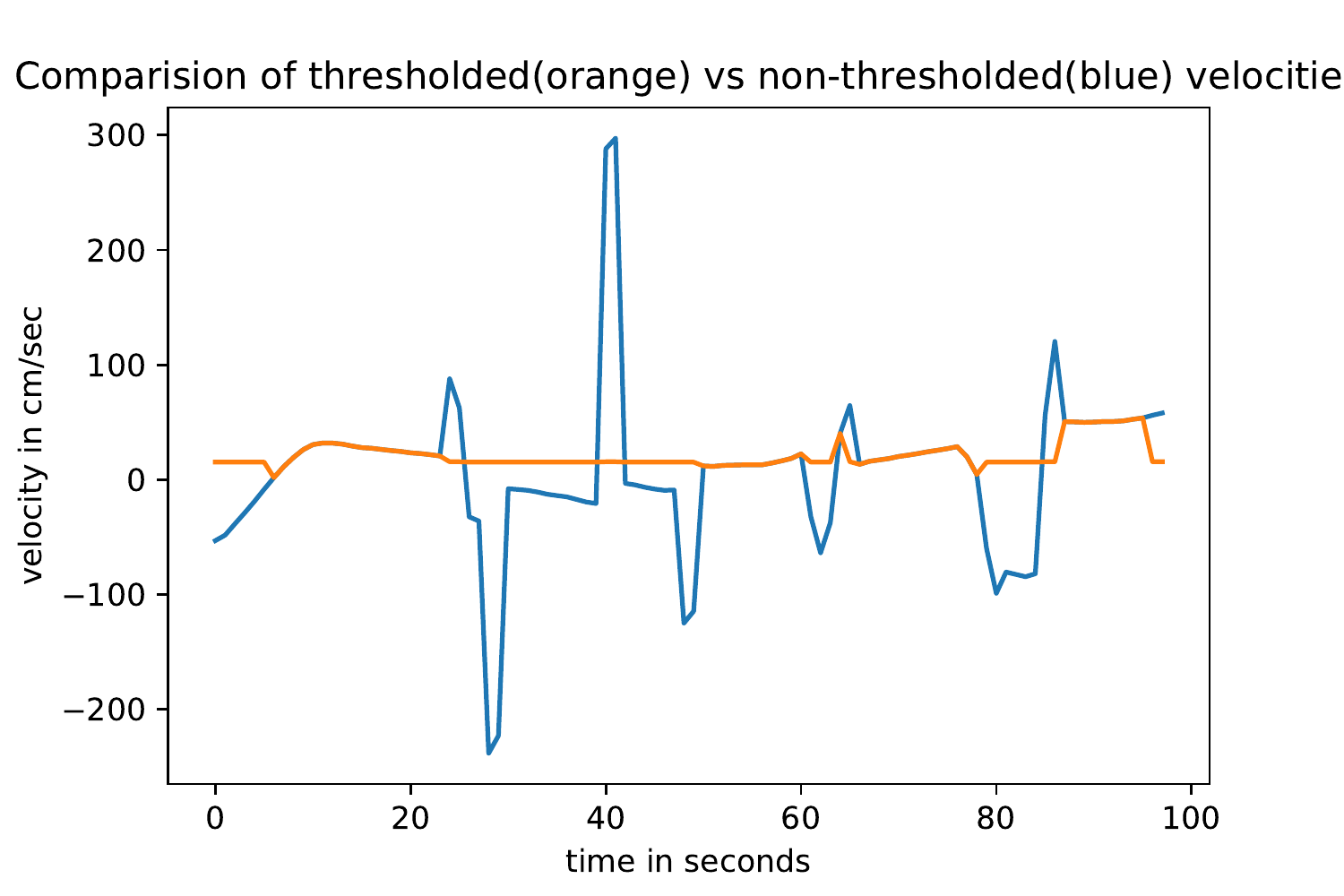}
\caption{Comparision of thresholded(orange) vs non-thresholded(blue) velocities}
\label{vel_comp}
\end{center}
\end{figure}

The thresholded velocities are replaced by median of velocities that are obtained. From the below figure \ref{vel_comp} we can observe the need for thresholding by looking at the unnatural velocities in detected because of not thresholding. we also can see the effect of thresholding

\subsection{Velocity Estimation}

\section{Experiments}
\subsection{Data and experimental setup}
Extensive video data was acquired using an IR MSA 5200HD2TIC Camera, with $N_x=320$ pixels and $\beta=49^o$. This camera is a multipurpose firefighting tool designed for search and rescue and structural firefighting. It uses an uncooled microbolometer vanadium oxide(Vox) detector which comprises of 320x240 FPA with the pitch of 38um and spatial resolution of 7.5 to 13.5um. The resolution is a necessary aspect to capture features for orientation estimation and IPS corresponding to this project. It records the image with 320x240
 focal plane array sensor. It has the ability to record imagery in two different modes, i.e. low and high  sensitivity modes. This device also features high score imagery generating 76000 pixels of image detail in low and high sense modes. Dense spectral resolution is (7.5 to 13.5 um) and the output video
 is in NTSC format with a frame rate of 30 fps and a scene temperature range of 560 degrees Celsius/
102 1040 degree Fahrenheit.

The dataset used in this project was recorded at Computer Engineering Building, University of New Mexico, Albuquerque and the Santa Fe Firefighting Facility, located in Santa Fe, New Mexico.

\subsection{Results}
\begin{figure}
\begin{center}
\includegraphics[width=0.4\textwidth]{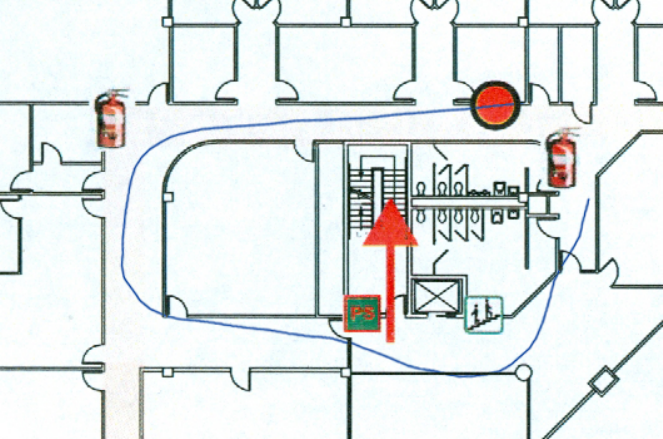}
\caption{Reconstruction of the path by our algorithm in exp-2}
\label{svr_con2}
\end{center}
\end{figure}

\begin{figure}
\begin{center}
\includegraphics[width=0.4\textwidth]{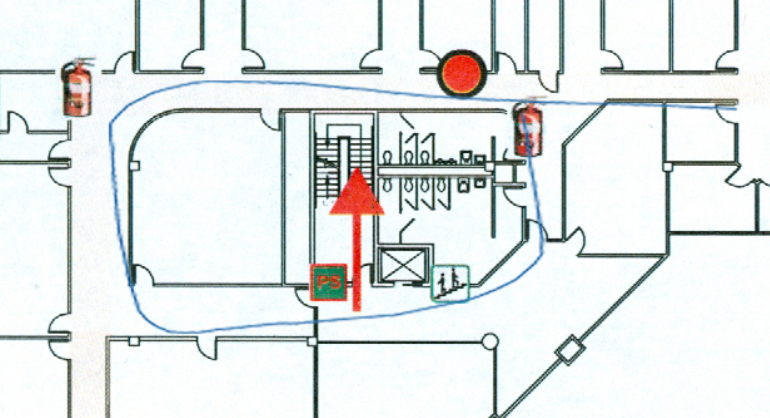}
\caption{Reconstruction of the path by our algorithm in exp-4}
\label{svr_con4}
\end{center}
\end{figure}

\begin{figure}
\begin{center}
\includegraphics[width=0.4\textwidth]{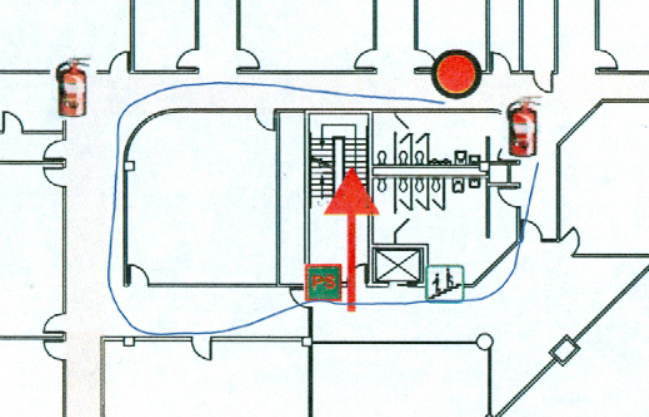}
\caption{Reconstruction of the path by our algorithm in exp-5}
\label{svr_con5}
\end{center}
\end{figure}

The video sequence was given as input to the IPS system for the reconstruction and obtained the path that was represented in blue solid line in the figures 7,8. In the below presented results \ref{svr_con2} \ref{svr_con4} \ref{svr_con5} we can see that the reconstruction is very accurate. Though there are probabilities for error the algorithm reconstructs the path with high accuracy $+- 2$ meters.

\section{Discussion and conclusion}
In the other prominent indoor positioning methods discussed in this paper they were able to achieve 2 meters accuracy \cite{1,3,5,6,7,21}. However, these require prior information regarding the scenario like positions of Bluetooth beacons and WiFi routers. In firefighting scenarios the prior information is unavailable. Hence the IPS algorithm we designed  will be highly reliable and recommended for unprepared firefighting scenarios.

We have implemented a novel IPS routine for firefighting scenarios. This technique is highly reliable and suitable for firefighter to get promising outcomes even if he is in an unprepared scenario.

 We would also like to implement reinforcement learning based real time route assistance by integrating the information from deep neural networks that were deployed to detect fire, rescue and augmented on the IPS map obtained.

\section*{Acknowledgements}
This work has been supported by NSF S\&CC EAGER grant 1637092.
\section*{References}
\bibliographystyle{IEEEtran}
\bibliography{citations}

\end{document}